\documentclass{llncs}
\usepackage{llncsdoc}
\usepackage{graphicx}
\usepackage{amssymb}
\usepackage{epstopdf}
\usepackage{color}
\usepackage{verbatim}
\usepackage[figuresright]{rotating}
\usepackage{algorithm2e}
\usepackage{multirow}
\def\bSig\mathbf{\Sigma}

\def\quad{ ~ }
\def\qquad{ ~~ }

\definecolor{darkgreen}{RGB}{0,190,0}       %This uses values 0 to 255
\begin{document}

%\thispagestyle{empty}

%\author{}
%\institute{}

%\title{Efficient Feature Selection for Large Target Feature Set based on Max-Relevance and Min-Redundancy}
\title{%TMRMR: Transductive Feature Selection based on Max-Relevance and Min-Redundancy for High Dimensional Data Regression Problems
MINT: Mutual Information based Transductive Feature Selection for Genetic Trait Prediction
}
\vspace{-.1in}
\author{Dan He\inst{1}, Irina Rish\inst{1}, David Haws\inst{1}, Simon Teyssedre\inst{2}, Zivan Karaman\inst{2}, Laxmi Parida\inst{1}}

\institute{IBM T.J. Watson Research,
Yorktown Heights, NY, USA\\
\{dhe, rish, dhaws, parida\}@us.ibm.com
\and
Limagrain Europe, Chappes Research Center,
CS 3911, 63720 Chappes, France\\ 
\{simon.teyssedre, zivan.karaman\}@limagrain.com
% For a paper whose authors are all at the same institution,
% omit the following lines up until the closing ``}''.
% Additional authors and addresses can be added with ``\and'',
% just like the second author.
%\and
}
\thispagestyle{empty}

\maketitle
%\section{}
%\subsection{}
%\vspace{-.3in}
%\begin{abstract}

%Nowadays it is common that in large-scale learning applications, the amount of variables far exceeds the number of samples. Various feature selection methods have been proposed for these applications to address the ``curse of dimensionality". In this work, we proposed a transductive feature selection method TMRMR based on maximum relevance and minimum redundancy. Our problem setting is the unlabeled test points are known to the method. TMRMR integrates the features for the test points into the selection process and our experiments show that the performance of TMRMR is much more significant than that of mRMR, the inductive version of the method. We also applied a dynamic programming algorithm for the greedy selection of the features and we show that our algorithm is of a few magnitude faster than mRMR.  Finally, as the greedy selection is usually not optimal, we applied a hill-climbing strategy to maximize the performance of TMRMR and our experiments show that the hill-climbing strategy is effective.

%\end{abstract}
\vspace{-.3in}
%\textbf{Keywords:} Feature Selection, Transductive, Relevance, Redundancy, Dynamic Programming

\section{Introduction}
%\vspace{-.1in}
%It is common that in large-scale learning applications, especially for biology data such as gene expression data, genotype data, the amount of variables far exceeds the number of samples. The ``curse of dimensionality" problem not only affects the computational efficiency of the learning algorithms, but also leads to poor performance of these algorithms. To address the problem, various feature selection methods have been proposed \cite{yang1997comparative}~\cite{guyon2003introduction}~\cite{jain1997feature}~\cite{dash1997feature} where a subset of important features are selected and the learning algorithms are trained only on these features.
\vspace{-.1in}
Whole genome prediction of complex phenotypic traits using high-density genotyping arrays has recently attracted a lot of attention, as it is relevant for the fields of plant and animal breeding and genetic epidemiology. Given a set of biallelic molecular markers, such as SNPs, with genotype values encoded as {0, 1, 2} on a collection of plant, animal or human samples, the goal is to predict the values of certain traits, usually highly polygenic and quantitative, by modeling simultaneously all marker effects, unlike the traditional GWAS.  rrBLUP
\cite{Meuwissen:2001fk,Whittaker:2000uq} has been used widely for trait prediction where it builds a linear model by fitting all the genotypes and the coefficient computed for each marker can be considered as a measure of the importance of the marker. The underlying hypothesis of normal distribution of marker effects is well suited for highly polygenic traits, and as the computations are fast and robust, it is one of the most used models in whole genome prediction. Other popular predictive models are Elastic-Net, Lasso, Ridge Regression\cite{Tibshirani94regressionshrinkage,Chen98atomicdecomposition} , Bayes A, Bayes B\cite{Meuwissen:2001fk}, Bayes C$\pi$\cite{kizilkaya2010genomic}, and Bayesian Lasso \cite{legarra2011improved,Park-Castella:2008}, etc.

As the number of genotypes is generally much bigger than the number of samples, the predictive models suffer from the ``curse of dimensionality". The ``curse of dimensionality" problem not only affects the computational efficiency of the learning algorithms, but can also lead to poor performance, mainly because of the correlation among markers. Feature selection \cite{yang1997comparative,guyon2003introduction,jain1997feature} has been considered as a successful solution for this problem, where a subset of important features are selected and the predictive models are trained only on these features. A popular criterion for feature selection is called \textit{Max-Relevance and
Min-Redundancy} (MRMR) \cite{peng2005feature} where the selected features are
maximally relevant to the class value and simultaneously minimally dependent on
each other. The method \textit{mRMR} \cite{peng2005feature} has been proposed
which greedily selects features that maximize the relevance while minimize
the redundancy. mRMR has been applied successfully on various
applications
\cite{huang2010analysis,cai2010prediction,zhang2008gene}.

\begin{comment}
To address the problem, various feature selection methods have been proposed
\cite{yang1997comparative,guyon2003introduction,jain1997feature} where a subset
of important features are selected and the learning algorithms are trained only
on these features.
\end{comment}

\begin{comment}
Predicting genetic traits from genotype data is a very important problem in
genetics and has a wide range of applications in heritability estimation, plant
breeding etc. Predictive models such as rrBLUP
\cite{Meuwissen:2001fk,Whittaker:2000uq}, Lasso \cite{Tibshirani94regressionshrinkage}, etc. have been applied where
the genotypes are treated as features and the values of the genetic traits are
treated as outputs. As the number of genotypes is generally very big, the
predictive models suffer from the ``curse of dimensionality''.  The ``curse of
dimensionality'' problem not only affects the computational efficiency of the
learning algorithms, but also leads to poor performance. To address the
problem, various feature selection methods have been proposed
\cite{yang1997comparative,guyon2003introduction,jain1997feature} where a subset
of important features are selected and the learning algorithms are trained only
on these features.
\end{comment}

Transductive learning, first introduced by Vapnik
\cite{gammerman1998learning}, assumes test data for predictor variables (here markers) are available to the learning algorithms (the target variable values for test samples are of course unknown). Therefore the models are built on both the training and test data, and usually lead to better predictive performance on the test data. In this work, we proposed a transductive feature selection method MINT based on information theory. MINT applies the MRMR criterion and integrates the test data in a natural way in the feature selection process. A dynamic programming algorithm is developed to speed up the selection process. Our experiments on both simulated and real data show that MINT generally achieves similar or better results than mRMR does which relies on training data only. To our knowledge, this is the first transductive feature selection method based on the MRMR criterion.

\vspace{-.1in}
\section{Methods}
\vspace{-.1in}
A popular criterion for feature selection is \textit{Max-Relevance and
Min-Redundancy} (MRMR) \cite{peng2005feature}. Max-Relevance searches for features
satisfying the Equation \ref{max_relevance}, which measures the mean value of
all mutual information values between individual feature $x_i$ and class
variable $c$.

\vspace{-.3in}
\begin{eqnarray}
max~D(S,c), D = \frac{1}{|S|} \sum_{x_i \in S} I(x_i;c)
\label{max_relevance}
\end{eqnarray}
%\vspace{-.2in}

\noindent where $S$ are the selected features, and $I(x_i;c)$ is the mutual information between $x_i$ and $c$.

However, feature selection just based on max-relevance tends to select features that have high redundancy, namely the correlations of the selected features tend to be big. If we remove some of the features that are highly correlated with other features, the respective class-discriminative power would not change much. Therefore, Min-Redundancy is proposed to select mutually exclusive features:

\vspace{-.2in}
\begin{eqnarray}
min~R(S), R = \frac{1}{|S|^2} \sum_{x_i, x_j \in S} I(x_i,x_j)
\label{min_redundancy}
\end{eqnarray}
%\vspace{-.2in}

An operator $\Phi(D,R)$ is defined to combine $D$ and $R$ from the above two equations where $D$ and $R$ are optimized at the same time:

\vspace{-.2in}
\begin{eqnarray}
max~\Phi(D,R), \Phi = D - R
\label{mrmr}
\end{eqnarray}
\vspace{-.2in}

In this work, based on the MRMR criterion, we proposed a novel method
\textbf{MINT} (\textbf{M}utual \textbf{IN}formation based \textbf{T}ransductive feature selection), which
targets feature selection with both the training data and the unlabeled test
data. We developed a dynamic programming based greedy
algorithm to efficiently select features.

\begin{comment}
An incremental search algorithm is applied to effectively find the near-optimal
features defined by $\Phi(.)$. The incremental algorithm works as the
following: Assuming feature set $S_{m-1}$ is already generated, which contains
$m-1$ features. The $m$-th feature needs to be selected from the set
${X-S_{m-1}}$, which maximizes the following objective function:

\vspace{-.2in}
\begin{eqnarray}
max_{x_j \in X-S_{m-1}} \big[I(x_j;c) - \frac{1}{m-1} \sum_{x_i \in S_{m-1}} I(x_j;x_i)\big]
\label{objective}
\end{eqnarray}
\vspace{-.2in}
\end{comment}

%\noindent The computational complexity of every single step in this algorithm is $O(|S| \times M)$ where $|S|$ is the size of the current target feature set, $M$ is the total number of features. Assuming the target feature set is eventually of size $N$, the complexity of this algorithm is $O(\sum_{i=1}^N i \times M)$ = $O(\frac{N^2}{2} \times M)$.
%\subsection{TMRMR(Transductive mRMR)}

%Transductive learning is shown to be able to improve the performance of the learning models solely trained on the training data. However, different learning methods require different adaptations of the transductive procedure and it is generally challenging to adapt the methods and it is not clear if all learning methods can be adapted.

We observe that the MRMR criterion has
two components, one for maximum relevance and one for minimum redundancy and that
the two components are independent. Maximum relevance requires calculation of
the mutual information between the selected features and the target variable. As in
transductive learning, the target variable values of the test samples are not available,
this component remains untouched. Minimum redundancy, on the other
hand, calculates the mutual information among all the selected features and the
target variable values are not involved. Therefore we can make the method transductive by
including all the test samples in this component to help improve the estimation of mutual information.

%Thus the objective function for MINT is thus defined as the following:
We applied the same incremental search strategy used in \cite{peng2005feature} to effectively find the near-optimal
features defined by $\Phi()$ in equation \ref{mrmr}. The incremental algorithm works as the
following: Assume feature set $S_{m-1}$ is already generated and contains
$m-1$ features. The $m$-th feature needs to be selected from the set
${X-S_{m-1}}$, which maximizes the following objective function:

\vspace{-.1in}
{\scriptsize
\begin{eqnarray}
 max_{x_j \in X-S_{m-1}} \big[I(x_j^{training};c^{training}) - \frac{1}{m-1} \sum_{x_i \in S_{m-1}} I(x_j^{training+test};x_i^{training+test})\big]
\label{objective_tmrmr}
\end{eqnarray}
}
\vspace{-.1in}

\noindent where $x_j^{training}$ denotes the $j$-th feature vector including only the training data, $x_j^{training+test}$ denotes the $j$-th feature vector including both the training and test data, $c^{training}$ denotes the class value vector including only the training data, $I(x_i, x_j)$ is the mutual information between $x_i$ and $x_j$. 

We next propose an efficient greedy algorithm to incrementally select the
features based on a dynamic programming strategy.  Our motivation is that the
operation $\sum_{x_i \in S_{m-1}} I(x_j;x_i)$ (for simplicity, we ignore the
superscripts of ``training" and ``test") need not be re-conducted for
every $x_j$. Since the features are added in an incremental manner, the
differences between $S_{m-1}$ and $S_{m-2}$ is just the $(m-1)$-th feature.
Therefore,  we do not need to re-compute the sum of the mutual information
between $x_j$ and $x_i$ where $1 \leq i \leq m-2$. The two sums
$\sum_{x_i \in S_{m-2}} I(x_j;x_i)$ and $\sum_{x_i \in S_{m-1}} I(x_j;x_i)$ are
just different by $I(x_j;x_{m-1})$. Therefore, we can save this sum $\sum_{x_i
\in S_{m-1}} I(x_j;x_i)$ at every step and reuse them in the next step. The
complexity of this dynamic programming algorithm is $O(NM)$, where $N$ is the
number of selected features and $M$ is the number of total features.

\vspace{-.2in}
\section{Experimental Results}
%\subsection{Simulated Data}
%We will show basically 3 cases by simulation: (1) mRMR = TMRMR $<$ rrBLUP, (2) rrBLUP $<$ mRMR = TMRMR, (3) rrBLUP $<$ mRMR $<$ TMRMR.
\vspace{-.1in}
We compare the predictive performance of rrBLUP \cite{Meuwissen:2001fk,Whittaker:2000uq} on the full set of variables versus its performance on the
subsets of variables of different size selected by  mRMR
\cite{peng2005feature} and MINT, referred to as ''mRMR + rrBLUP'' and ''MINT + rrBLUP'', respectively.
%As both mRMR and MINT are feature selection methods, we apply them first to select a set of features and then run rrBLUP on these selected features.
Similar results, not included due to space restrictions, were obtained when  applying some  other predictive methods to features selected by mRMR and MINT.  
In all experiments, we perform  10-fold cross-validations and measure the average coefficient of determination $r^2$ (computed as the square of Pearson's correlation coefficient) between
the true and predicted outputs;  higher $r^2$ indicates better performance.  %We repeat the experiments for each parameter settings ten times and give the average results.
\vspace{-.1in}
\paragraph{\bf Simulated Data.}
As our method is based on the MRMR criterion, we experiment with different levels of 
 relevance and redundancy, and show that the performance of MINT
relies on both components. We randomly simulate 10 different data sets  for each parameter settings, and report  average results.
\begin{comment}
We first simulate a target vector variable $Y$ with random values in [0,1] with
dimension 200, namely we consider 200 samples. For the design matrix $X$, we
simulated 100 ``good" features by adding an error vector of the same dimension
200 as $e \sim N(0, 300)$ to the $Y$ vector. We also simulate 1900 ``bad"
features by adding an error vector of dimension 200 as $e' \sim N(0, 500)$.
Therefore it is relatively hard to distinguish the good and bad features. The
results are shown in Table \ref{simulation}, Case one. In this case the feature
selection methods do not work well as it's hard to select good features.
\end{comment}
First, we simulate a 200-dimensional target variable vector $Y$ following  multivariate uniform distribution $Y \sim U(0,1)$, and then we simulate the features as $F = Y + e$, where $e \sim N(0, \delta^2)$ is a 200-dimensional noise vector  following a multivariate normal distribution. Thus, the large-$\delta$ features are noisy (``bad''), while lower-$\delta$ features are less noisy (``good'').
 %as ''bad contain more noise and are relatively ``bad" in a regression model. %On the contrary, with smaller $\delta$, the features contain less noise and are relatively ``good".
We simulate 100 ``good" features with $e \sim N(0,100)$ and 1900 ``bad" features with $e' \sim N(0, 1000)$. % Obviously the good and bad features are relatively easy to distinguish. 
The results are shown in
Table \ref{simulation}, Case one. In this case the feature
selection methods work well, but as the good features are randomly simulated
and they have low redundancy, the performances of mRMR+rrBLUP and MINT+rrBLUP are almost
identical.

\begin{comment}
For the design matrix $X$, we simulate 100
``good" features by adding an error vector of dimension 200 as $e \sim N(0,
100)$ to the $Y$ vector. We also simulate 1900 ``bad" features by adding an
error vector of dimension 200 as $e' \sim N(0, 1000)$. Therefore the good and
bad features are relatively easy to distinguished. The results are shown in
Table \ref{simulation}, Case two. In this case the feature
selection methods work well, but as the good features are randomly simulated
and they have low redundancy, the performances of mRMR and MINT are almost
identical.
\end{comment}

Next, we again simulate a target variable vector $Y \sim U(0,1)$. For the design matrix $X$, we simulate 50 ``seed" features $F = Y + e$ with $e \sim N(0, 500)$. Then for each seed feature, we simulate 9 ``duplicate" features as $F' = F + e'$ where $e' \sim N(0, 100)$. We consider all these 500 features as ``good" features. We also simulate 4500 ``bad" features
$F'' = Y + e''$ with $e'' \sim N(0, 1000)$. Therefore
the good and bad features are still relatively easy to be distinguished and there are
large redundancies among the good features. The results are shown in Table
\ref{simulation}, Case two.  MINT+rrBLUP  consistently
outperforms  mRMR+rrBLUP, due to the redundancy we introduced in
the good feature set; both methods   outperform  rrBLUP.

\vspace{-.3in}
\begin{table}[ht]
\caption{Performance (average $r^2$ over 10-fold CV) of rrBLUP on the full set of features  vs. MINT+rrBLUP and mRMR+rrBLUP,  
on simulated data for two different cases.}
\centering
\begin{tabular}{|c|c|c|c|c|} \hline
Case &	 rrBLUP & Number of   & MINT & mRMR \\
     &   (all features) & selected features  & + rrBLUP & +rrBLUP \\ \hline
%\multirow{2}{*}{One} & 50 & \multirow{2}{*}{0.91}   & 0.641 & 0.639 \\
%&100 &   & 0.725 & 0.73 \\\hline
\multirow{2}{*}{One} & \multirow{2}{*}{0.845}  & 50   & 0.940 & 0.938 \\
& & 100  & 0.958 & 0.962 \\\hline
\multirow{5}{*}{Two} &  \multirow{5}{*}{0.187} &150  & 0.281 & 0.280 \\
& &250 &     0.376 & 0.363 \\
& &350 &   0.434 & 0.411 \\
& &450 &    0.432 &  0.414 \\
& &550 &    0.456 & 0.448 \\\hline
\end{tabular}
\vspace{-.3in}
\label{simulation}
\end{table}

%\vspace{-.3in}
%\subsection{Real Data}
%\vspace{-.1in}
%We compare the performance of rrBLUP \cite{}, mRMR \cite{} and MINT on the two Maize data sets \cite{} Dent and Flint. As each data set has three phenotypes, we in total have 6 data sets. We vary the number of selected features. As both mRMR and MINT are feature selection methods, we apply them first to select a set of features and then run rrBLUP on these selected features.

\paragraph{\bf Real data.}
%We compare the performance of rrBLUP without any feature selection vs. its performance on the features selected by mRMR and MINT
Next, we compare the same methods  on the two Maize data sets
Dent and Flint \cite{rincent2012maximizing} and show the results in Tables \ref{dent_flint}. Each data set, Dent and Flint, has three phenotypes,
thus we have six phenotypes overall. Dent has 216 samples and 30,027 features
and Flint has 216 samples and 29,094 features.  We vary the number of selected
features as 100, 200, 300, 400 and 500. It is obvious that both mRMR+rrBLUP and MINT+rrBLUP
outperform rrBLUP significantly, indicating feature selection in general is
able to improve the performance of the predictive model. On the other hand, in
almost all data sets, MINT outperforms mRMR consistently, illustrating the
effectiveness of transduction.

%We will also show the performance of TMRMR on one more data set.
\vspace{-.2in}
\begin{table}[ht]
\caption{Performance (average $r^2$ over 10-fold CV) of rrBLUP (all features)  vs. mRMR+rrBLUP  and MINT+rrBLUP  on Maize (Dent and Flint) and $n$ is the number of selected features.}
\centering
\begin{tabular}{|l|c|c|c|c|c|c|c|} \hline
	Data Set (\textbf{Dent}) &  rrBLUP (all features) & Heritability & n=100 & n=200 & n=300 & n=400 & n=500  \\\hline
Pheno 1 (mRMR+rrBLUP) & \multirow{2}{*}{0.439} &   \multirow{2}{*}{0.952} & 0.426 & 0.514 & 0.526 & 0.563 & 0.536  \\
Pheno 1 (MINT+rrBLUP) &  &  & 0.623 & 0.662 & 0.668 & 0.653 & 0.663\\\hline
Pheno 2 (mRMR+rrBLUP) &  \multirow{2}{*}{0.410} &  \multirow{2}{*}{0.932}& 0.584  & 0.591 & 0.603 & 0.619 & 0.629  \\
Pheno 2 (MINT+rrBLUP) &   &  & 0.687 & 0.674 & 0.678 & 0.667 & 0.669\\\hline
Pheno 3 (mRMR+rrBLUP) & \multirow{2}{*}{0.228} & \multirow{2}{*}{0.791} & 0.502 & 0.523 & 0.521 & 0.515 & 0.498 \\
Pheno 3 (MINT+rrBLUP) &  &  & 0.517 & 0.515 & 0.514 & 0.536 & 0.537\\\hline
	Data Set (\textbf{Flint}) &  rrBLUP (all features) & Heritability & n=100 & n=200 & n=300 & n=400 & n=500\\
\hline
Pheno 1 (mRMR+rrBLUP) & \multirow{2}{*}{0.275} & \multirow{2}{*}{0.954} &  0.493 & 0.472 & 0.476 & 0.495 & 0.508 \\
Pheno 1 (MINT+rrBLUP) &  &  & 0.627 & 0.599 & 0.606 & 0.588 & 0.601\\\hline
Pheno 2 (mRMR+rrBLUP) &  \multirow{2}{*}{0.255} & \multirow{2}{*}{0.643} & 0.340  & 0.407 & 0.411 & 0.408 & 0.405 \\
Pheno 2 (MINT+rrBLUP) &   &  & 0.480 & 0.500 & 0.498 & 0.486 & 0.491\\\hline
Pheno 3 (mRMR+rrBLUP) & \multirow{2}{*}{0.047} & \multirow{2}{*}{0.355} & 0.199 & 0.250 & 0.288 & 0.275 & 0.271  \\
Pheno 3 (MINT+rrBLUP) &  &  & 0.320 & 0.350 & 0.361 & 0.352 & 0.350\\\hline
\end{tabular}
\vspace{-.2in}
\label{dent_flint}
\end{table}

\begin{comment}
\begin{table}[ht]
\caption{Performance (average $r^2$ over 10-fold CV) of rrBLUP (all features) vs. mRMR+rrBLUP and MINT+rrBLUP on Maize (Flint).}
\centering
\begin{tabular}{|l|c|c|c|c|c|c|} \hline
	Data Set &  rrBLUP (all features) & n=100 & n=200 & n=300 & n=400 & n=500\\
\hline
Pheno 1 (mRMR+rrBLUP) & \multirow{2}{*}{0.275} &  0.493 & 0.472 & 0.476 & 0.495 & 0.508\\
Pheno 1 (MINT+rrBLUP) &  &  0.627 & 0.599 & 0.606 & 0.588 & 0.601\\\hline
Pheno 2 (mRMR+rrBLUP) &  \multirow{2}{*}{0.255} & 0.340  & 0.407 & 0.411 & 0.408 & 0.405\\
Pheno 2 (MINT+rrBLUP) &   &  0.480 & 0.500 & 0.498 & 0.486 & 0.491\\\hline
Pheno 3 (mRMR+rrBLUP) & \multirow{2}{*}{0.047} & 0.199 & 0.250 & 0.288 & 0.275 & 0.271\\
Pheno 3 (MINT+rrBLUP) &  & 0.320 & 0.350 & 0.361 & 0.352 & 0.350\\\hline
\end{tabular}
\vspace{-.2in}
\label{flint}
\end{table}
\end{comment}

\begin{comment}
\begin{table}[ht]
\caption{Performance of rrBLUP VS. MRMR and TMRMR on Rice }
\centering
\begin{tabular}{|c|c|c|c|c|c|c|} \hline
	Data Set &  rrBLUP & n=100 & n=200 & n=300 & n=400 & n=500\\\hline
29 (mRMR) & 0.275121 &  0.3247263 & 0.3575819 & 0.3701481 & 0.3676283 & 0.3588806\\
29 (TMRMR) & 0.275121 &  0.3247263 & 0.354183 & 0.6059768 & 0.3648831 & 0.3608316\\
34 (mRMR) &  0.2553844 & 0.1609234  & 0.1948684 & 0.193073 & 0.2020893 & 0.2123053\\
34 (TMRMR) &  0.2553844 &  0.1599493 & 0.1934766 & 0.4979312 & 0.2033646 & 0.2110646\\\hline
\end{tabular}
\vspace{-.2in}
\label{epistasis}
\end{table}
\end{comment}

\begin{table}[ht]
\caption{Performance of Feature Selection}
\centering
\begin{tabular}{|c|c|c|c|c|c|c|c|} \hline
	Data Set & MINT+rrBLUP (n=500) & mRMR+rrBLUP (n=500) & rrBLUP  & Lasso & Elastic Net & SVR & num. of samples \\\hline
	Data 1 & \textbf{0.14} & 0.015 & 0.018 & 0.018 & 0.019 & 0.079 & 220  \\\hline
	Data 2 & \textbf{0.303} & 0.110 & 0.204 & 0.129 & 0.146 & 0.234 & 319 \\\hline
	Data 3 & 0.563 &  0.536 & 0.607 & 0.540 & 0.545 & \textbf{0.615} & 1217  \\\hline
	Data 4 & \textbf{0.369} & 0.235 & 0.31 & 0.255 & 0.261 & 0.306 & 671  \\\hline
	Data 5 & \textbf{0.364} & 0.201 &  0.22 & 0.212 & 0.214 & 0.174 & 532  \\\hline
	Data 6 & \textbf{0.276} & 0.148 &  0.17 & 0.15 & 0.162 & 0.184 & 620  \\\hline
	%k = 5 & 0.089 &  & & & \\\hline
\end{tabular}
\label{FS}
\end{table}

\vspace{-.3in}
\section{Conclusions}
\vspace{-.1in}
In this work, we proposed a transductive feature selection method MINT based on
information theory where the test data is integrated in a natural way into a
greedy feature selection process. A dynamic programming algorithm is developed
to speed up the greedy selection. Our experiments on both simulated and real
data show that MINT is generally a better method than the inductive feature
selection method mRMR. What's more, MINT is not restricted to genetic trait prediction problems but is a generic feature selection model. 

\vspace{-.2in}
\bibliographystyle{plain}
\begin{small}
\bibliography{mRMR}
\end{small}

\end{document}